\documentclass[runningheads]{llncs}
\usepackage{eccv}
\usepackage{eccvabbrv}
\usepackage{multirow}
\usepackage{graphicx}
\usepackage{booktabs}

\usepackage{array}
\newcommand{\PreserveBackslash}[1]{\let\temp=\\#1\let\\=\temp}
\newcolumntype{C}[1]{>{\PreserveBackslash\centering}p{#1}}
\newcolumntype{R}[1]{>{\PreserveBackslash\raggedleft}p{#1}}
\newcolumntype{L}[1]{>{\PreserveBackslash\raggedright}p{#1}}

\usepackage[misc]{ifsym}
\usepackage[accsupp]{axessibility}  
\usepackage{hyperref}
\usepackage{orcidlink}

\begin{document}
\title{\textcolor{blue}{\textsf{Turbo}}: Informativity-Driven Acceleration \\ Plug-In for Vision-Language Large Models}
\titlerunning{Turbo}

\author{Chen Ju\inst{1*} \and
Haicheng Wang\inst{1,2*} \and
Haozhe Cheng\inst{2} \and
Xu Chen\inst{1} \and
Zhonghua Zhai\inst{1} \and
Weilin Huang\inst{1} \and
Jinsong Lan\inst{1} \and
Shuai Xiao\inst{1}\,\textsuperscript{\Letter} \and
Bo Zheng\inst{1}}

\authorrunning{C. Ju et al.}
\institute{$^1$Alibaba Group, \quad $^2$Shanghai Jiao Tong University \\
\email{cju.void@gmail.com, anakin\_skywalker@sjtu.edu.cn, shuai.xsh@alibaba-inc.com}
\\
\url{https://voide1220.github.io/turbo/}
}

\maketitle

\begingroup
\renewcommand\thefootnote{\relax}\footnotetext{*: These authors contribute equally to this work.}
\endgroup

\begin{abstract}
Vision-Language Large Models (VLMs) recently become primary backbone of AI, due to the impressive performance. However, their expensive computation costs, {\em i.e.}, throughput and delay, impede potentials in the real-world scenarios. To achieve acceleration for VLMs, most existing methods focus on the model perspective: pruning, distillation, quantization, but completely overlook the data-perspective redundancy. To fill the overlook, this paper pioneers the severity of data redundancy, and designs one plug-and-play Turbo module guided by information degree to prune inefficient tokens from visual or textual data. In pursuit of efficiency-performance trade-offs, information degree takes two crucial factors into consideration: mutual redundancy and semantic value. Concretely, the former evaluates data duplication between sequential tokens; while the latter evaluates each token by its contribution to the overall semantics. As a result, tokens with high information degree carry less redundancy and stronger semantics. For VLMs' calculation, Turbo works as a user-friendly plug-in that sorts data referring to information degree, utilizing only top-level ones to save costs. Its advantages are multifaceted, {\em e.g.}, being generally compatible to various VLMs across understanding and generation, simple use without re-training and trivial engineering efforts. On multiple VLMs benchmarks, we fully experiment to demonstrate the good acceleration of Turbo, under negligible performance drop. 
\end{abstract}

\section{Introduction}  \label{sec:intro}
Vision-Language Large Models (VLMs)~\cite{Radford21,Jia21,li2022blip,li2023blip,kim2021vilt} have achieved promising progress towards AI. Inspired by superior performance and emergent abilities, VLMs are even considered as one of the future trend towards AGI. As multiple studies~\cite{wei2022emergent,brown2020language} illustrated, VLMs' capability is closely linked to: model parameters and data quality. Therefore, expanding model scale and feeding high-quality data to improve practical performance, has become the community consensus. As expected, VLMs show superior results on various downstream tasks/domains~\cite{rombach2022high,ju2022prompting,zhou2022learning}, including understanding and generation, with delicate architectures.

\begin{figure}[t]
\begin{center}
\includegraphics[width=1\textwidth] {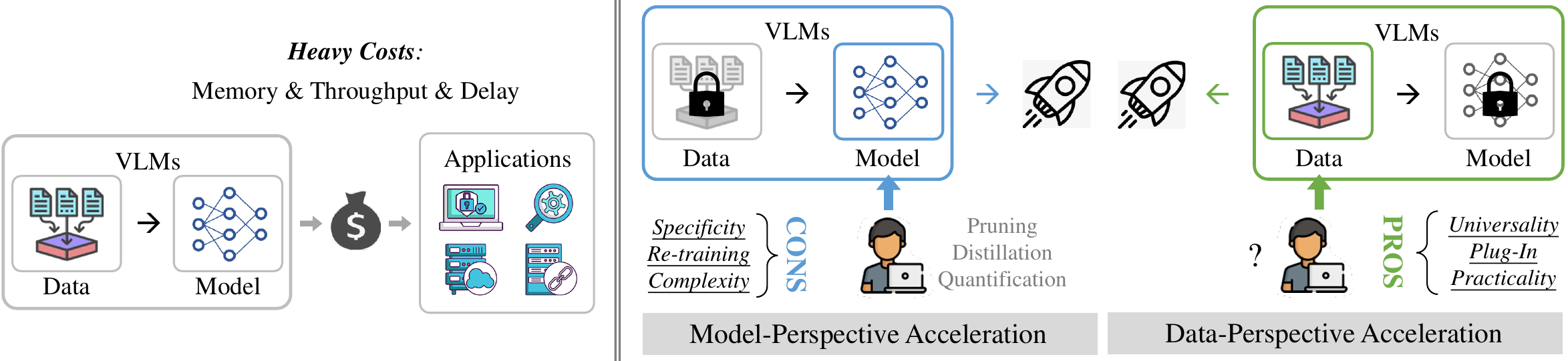}
\end{center}
\caption{\textbf{Left}: the trouble with applying VLMs is the high-cost issue. \  \textbf{Right}: to accelerate VLMs, most existing ideas focus on the model perspective (pruning \& quantization). While our Turbo explores de-redundancy from the data perspective.}
\label{fig:intro}
\end{figure}

Although it seems that building leading to AI has been completed, there is still a dark cloud hanging over VLMs, making everything a utopia. That is, the cost issue (computational throughput/latency/memory), whether it is training or deployment. Let's imagine, for text-to-image generation, the snail's pace of one picture every 10 seconds is one bottleneck for real-world applications. As a result, reducing costs becomes key to promoting the popularity of VLMs.

To achieve acceleration for VLMs, existing methods~\cite{wang2022efficientvlm,shen2020q,frantar2022gptq,liu2023deja} focus on one \textbf{Model-Center Perspective (MCP)}, {\em e.g.}, pruning, distillation, and quantization, as shown in Figure~\ref{fig:intro}. Although effective, they suffer from tricky issues: \underline{\textit{Specificity}}: MCP is only developed for specific architectures with poor generalization, implying it's incompatible with various VLMs, {\em e.g.}, MCP acceleration suitable for understanding is infeasible for generation. \underline{\textit{Re-training}}: To maintain high performance, MCP usually requires re-training or fine-tuning VLMs, which inevitably consumes additional computing overheads, being inefficient to apply. \underline{\textit{Complexity}}. The development of MCP involves considerable tricks, which raise barriers to applications. To sum up, with only efforts on the model side, MCP is trapped in a dilemma: still high costs, and poor generalization.

To jump out of the dilemma, as AI systems typically cover ``data \& model'', this paper raises one novel question about acceleration, from one \textbf{Data-Center Perspective (DCP)}: \textit{Is there redundancy in the data side? And if so, how high?} To answer the question, we first define, then evaluate the informativity of token sequences in each VLMs block. The exploration motivation is that attention-based networks have emerged as a dominant architecture among VLMs, resulting in a quadratic relationship between computation and input sequence length. The results show that, token redundancy is consistently high in most blocks. Besides, token redundancy gradually increases as blocks deepen. With these observations, we conclude that acceleration by data de-redundancy is promising.

To enable data acceleration, we propose one novel Turbo plug-in, with the spirit that compressing invalid components while retaining semantic essences. Concretely, we define an information degree, covering two components: mutual redundancy and semantic value. The former evaluates the information duplication between sequential tokens, while the latter focuses on the token’s contribution to sample-wise semantics. For mutual redundancy, the insight is that tokens with dependency tend to have similar informativity, making re-use feasible. For semantic value, the insight is that tokens with core contributions maintain strong performance as principal components. Using information degree of sequential tokens, Turbo naturally sorts data to only leverage the top-level ones for VLMs' calculation, thus saving costs from the source. To sum up, Turbo wins considerable pros. \underline{\textit{Universality}}. The accelerated data is compatible with various VLMs, {\em e.g.}, understanding/generation, multi-/uni-modality, showing the powerful generalization. \underline{\textit{Plug-and-Play.}} Turbo involves no training parameters, which is lightweight to avoid trivial development efforts. \underline{\textit{Practicality}}. Turbo is user-friendly with no need for tedious tricks. And even more valuable, it could superpose on existing model-perspective acceleration.

Under two types of VLMs, namely, image-align-text understanding and text-to-image generation, we experiment to prove the acceleration generality of Turbo, across several standard datasets. For almost all understanding tasks (retrieval, classification, caption and VQA), Turbo improves throughput by around $2$X with little loss of performance. For most generation tasks (text-to-image and image-to-image), Turbo improves throughput to $1.6$X without compromising quality. We also ablate the component effectiveness, both quantitatively and qualitatively.

To sum up, our contributions lie in three folds:

$\bullet$ We explore a data-perspective acceleration for VLMs. It's generally adapted to understanding \& generation, is orthogonal to model-perspective acceleration.  

$\bullet$ We design one novel Turbo for data de-redundancy, by evaluating information degree of redundancy and semantics. Turbo serves as training-free plug-in for VLMs to achieve great trade-offs between efficiency-performance. 

$\bullet$ We conduct extensive experiments and thorough ablations to reveal good acceleration of Turbo, and our superior results for understanding/generation.

\section{Related Work}  \label{sec:related work}
\noindent \textbf{Vision-Align-Language Understanding} gets image-text shared embeddings, by pre-training with large-scale data. Recent studies mainly divided into: single tower~\cite{chen2020uniter,kim2021vilt}, twin towers~\cite{Radford21,Jia21}, and bridge tower~\cite{li2023blip,li2022blip,zhu2023minigpt}. Structurally, attention-based Transformer has emerged as a dominant architecture, greatly increasing computing overhead while improving performance. They have brought many promising potentials in terms of understanding videos~\cite{ju2021divide,ju2022adaptive,ju2020point}, audio~\cite{liu2024annotation,liu2023audio,liu2024audio} and image~\cite{cheng2023mixer,chen2024enhancing,cheng2023image,ye2021unsupervised}. They are broken down into: recognition~\cite{ju2023distilling,cheng2024denoiser,zhao2020bottom}, grounding~\cite{ju2023constraint,liu2022exploiting}, segmentation~\cite{ma2023open,yang2023multi,ma2023attrseg}, caption~\cite{mokady2021clipcap,luo2022clip4clip}, retrieval~\cite{ju2022prompting,song2022clip}.

\noindent \textbf{Vision-Language Generation} aims to build compositional models from text \& noise to pixel-wise vision. Typically, research lines are mainly divided into: Diffusion (SD)~\cite{ho2020denoising,rombach2022high} and DALL-E~\cite{ramesh2022hierarchical,shi2020improving}. Although they have shown promising potentials~\cite{ma2023diffusionseg,ju2023multi,chen2024wear}, expensive computing costs seriously damage user experience. For an instance, when using SD for one $1024*1024$ image, the inference delay is about $8$ seconds. Structurally, SD’s UNet accounts for most overhead.

\noindent {\bf Acceleration for VLMs.}
To alleviate high computing costs in the uni-modal transformer-based models, extensive acceleration studies has been proposed~\cite{frantar2022gptq,rao2021dynamicvit,wu2023ppt,fayyaz2022adaptive,wei2023joint,xiao2023smoothquant,xu2022evo,liang2022not,liu2023simple,bolya2023token}. However, such an exploration on VLMs is still in its infancy, mainly because of the difficulty to unify different modalities under one acceleration paradigm. Nowadays, most solutions for VLMs acceleration are from the model-perspective: knowledge distillation~\cite{wang2022efficientvlm,fang2021compressing}, floating quantization~\cite{shen2020q,frantar2022gptq}, and model pruning~\cite{wang2022efficientvlm,shi2023upop,jiang2022trips}. Although achieving high sparsity with competitive performance, they all suffer from multiple issues. \textit{Re-training}. To maintain high performance, fine-tuning VLMs is usually inevitable, which is inefficient to bring additional computing burdens. \textit{Specificity}. They are generally developed for specific networks, which have poor generalization to compatible with various VLMs. \textit{Complexity}. They can involve many empirical tricks, these trivial matters raises barriers to application. Overall, in contrast, Turbo is totally training-free, can be easily plugged in most VLMs with competitive performance.

Recently, ToMe~\cite{bolya2022token} is designed as an acceleration plug-in for vision transformers, but Turbo differs from it in two aspects. \textit{Universality:} ToMe is designed for uni-modal image classification; while Turbo is generally compatible to various VLMs: understanding/generation, multi-/uni-modality. \textit{All-round:} ToMe focuses solely on mutual redundancy; while our Turbo combines mutual redundancy and semantic value, to achieve better acceleration-performance trade-offs.

\begin{figure*}[t]
\begin{center}
\includegraphics[width=1\textwidth] {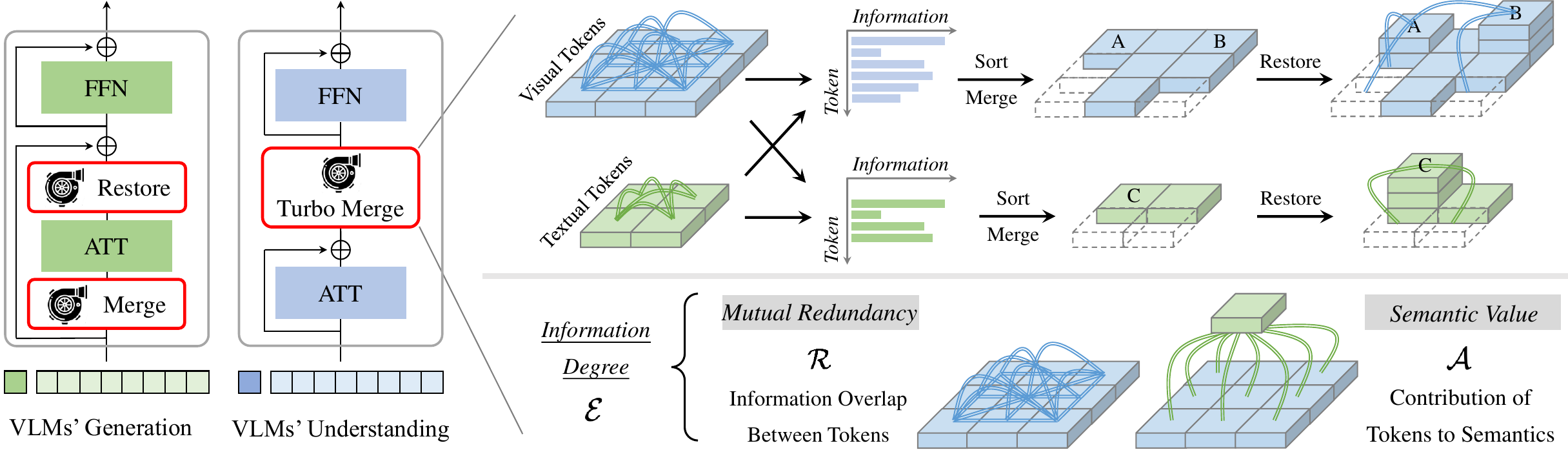}
\end{center}
\caption{\textbf{Computing Architecture.} As one plug-in module, Turbo compresses data to cut computing overheads for various VLMs, across understanding/generation and uni-/multi-modality. It sorts then merges tokens by information degree (mutual redundancy $\mathcal{R}$ and semantic value $\mathcal{A}$) for understanding tasks; while sorts, merges and restores VLMs’ tokens for generation tasks, owning good universality and practicality.}
\label{fig:framework}
\end{figure*}

\section{Methods: Acceleration for VLMs}   \label{method} 
To alleviate heavy deployment costs from VLMs, we conduct thorough analysis in terms of data; then describe informativity-driven turbo plug-in for acceleration.

\noindent {\bf Revisit of Attention.} \ \ 
For VLMs, Transformer composed of attention has emerged as a dominant architecture. For either text or image data, we generally processes it as the $n$-token 1D sequence $\mathbf{X} \in \mathbb{R}^{n \times D}$, then pass into the attention layer to model sequence relationships globally, where one weighted sum of values based on the affinity over other tokens is calculated. Attention is formulated as: 
\begin{equation} \label{eq:att}
    {\mathrm{Attention}(\mathbf{Q,K,V}) = \mathrm{softmax}(\frac{\mathbf{Q}\mathbf{K}^T}{\sqrt{D}})\mathbf{V}.}
\end{equation}

Transformer achieves impressive performance, comparing to previous architectures, {\em e.g.}, convolution and graph, but at the cost of being more expensive. It is because the complexity of multi-head attention modules (self-attention or cross-attention) scales quadratically with the sequence length $n$. And thus, long tokens lead to substantial computational overheads.

With the fundamental understanding, we raise one novel question: \textit{Is there any redundancy in the token sequence? If so, we could remove such redundancy from one data perspective, to cut off computational overheads from the source.}

\subsection{Multi-Modal Informativity} \label{sec:theory} 
\noindent {\bf Motivation \& Insight.} Given that most input data are likely to contain superfluous parts, {\em e.g.}, image background, irrelevant objects, we foresee a significant opportunity to further ``compress'' data. To accelerate VLMs from the data perspective, the primary process is to distinguish where the redundancy lies. And to analyse it quantitatively, we define token sequence informativity as a measure to quantify the information contained in one token sequence.

Inspired by the informativity idea in information theory, we define $(\Omega, \mathcal{F}, \mathbb{P})$ as the probability space of token sequences, where the triplet is sample space, $\sigma$-algebra, and probability measure. Following the concept of self-information~\cite{shannon1948mathematical}, we define the informativity of one token sequence $\mathbf{X}=[x_1,x_2,...,x_n]$ as:
\begin{equation}
{\mathcal{I}(\mathbf{X})= -\mathrm{log} \, \mathbb{P}(\mathbf{X}) = -\mathrm{log} \, \mathbb{P}([x_1,x_2,...,x_n]).}
\end{equation}
By the compound probability formula, we have
\begin{equation}
\mathcal{I}(\mathbf{X}) = -\mathrm{log} \, \mathbb{P}(x_{k, k\in\psi} | x_{i,i\in \{\phi -\psi\}}) \cdot \mathbb{P}(x_{i,i\in \{\phi -\psi\}}),
\end{equation}
where $\phi = \{1,...,n\}$ and $\psi$ is a subset of $\phi$. If we can find $\psi$ such that
\begin{equation}
\mathbb{P}(x_{k, k\in\psi} | x_{i,i\in \{\phi -\psi\}}) \approx 1,
\end{equation}
then the informativity of the token sequence can be approximated by
\begin{equation}
\mathcal{I}(\mathbf{X}) \approx -\mathrm{log} \, \mathbb{P}(x_{i,i\in \{\phi -\psi\}}).
\end{equation}
As revealed by MAE series~\cite{he2022masked,tong2022videomae,huang2024mavil}, there exists considerable redundancies in visual tokens, such that only $25$\% tokens almost restore the whole sequence, validating the existence of subset $\psi$. And hence, the subset $\{x_{k, k\in\psi}\}$, having a total dependency on its complementary set $\{x_{i,i\in \{\phi -\psi\}}\}$, enables us to aggregate $\{x_{k, k\in\psi}\}$ into $\{x_{i,i\in \{\phi -\psi\}}\}$ for efficient calculation with little information loss.

\noindent {\bf Mutual Redundancy.} Based on the analysis, we aim to find the dependencies between tokens of the given sequence, so that helping discover subset $\psi$.

One intuition is, tokens with mutual dependency tend to have similar representations, {\em e.g.}, we can easily restore the texture of a shirt by just a few patches from it. What's more, tokens with high similarity have analogical contributions in the attention calculation. Hence, we believe that by merging tokens with high mutual similarity, $\{x_{k, k\in\psi}\}$ with a high dependency on $\{x_{i,i\in \{\phi -\psi\}}\}$ can be constructed. Formally, for token $x_i$, we define its mutual redundancy $\mathcal{R}_i$ to be: 
\begin{equation}
    \mathcal{R}_i = \mathrm{Max}
    \{\mathcal{S}(x_i, x_j), \ \,  j \in \{1,...,n\}\backslash i\},  
\end{equation}
where $\mathcal{S}(\cdot,\cdot)$ refers to cosine similarity, and $\mathrm{Max}$ is the maximum operation.

\begin{figure}[t]
\begin{center}
\includegraphics[width=1\textwidth] {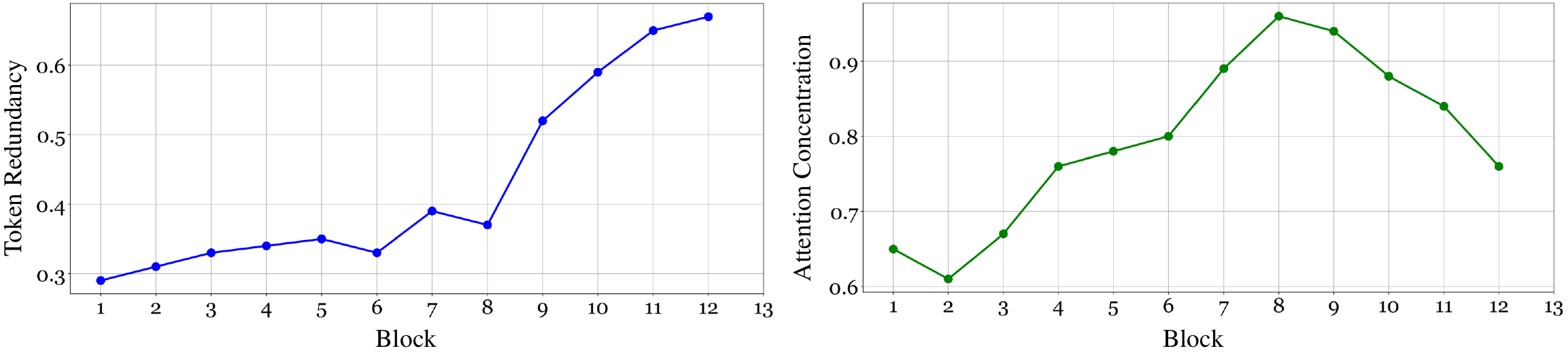}
\end{center}
\caption{\textbf{Empirical Evaluation of Token Redundancy \& Attention Concentration} on BLIP fine-tuned for multi-modal retrieval. Results reveal the non-negligible redundancy in the token sequence from perspectives of semantics and similarity.}
\label{fig:evaluation}
\end{figure}

\noindent {\bf Semantic Value.} \ 
Mutual redundancy for data merging could already get relatively good results on simple tasks, {\em i.e.}, image classification. However, in this process, each token is treated equally, thus its contribution to semantic categories are neglected. This will result in an early merging for tokens with important semantics, causing a drastic performance drop on fine-grained, information-demanding tasks, {\em i.e.}, multi-modal understanding and cross-modal generation.

To merge tokens with preferences from trivial background to significant foreground, we use the guidance of semantic value hidden inside network. By defining $\mathbf{Y}$ to be the $\mathit{cls}$ token of the output sequence $\mathbf{X}_{out}$ from $\mathbf{X}$ through the attention block, We treat $\mathbf{Y}$ as rich semantic guidance, calculate its informativity as:
\begin{equation}
\mathcal{I}(\mathbf{Y})= -\mathrm{log} \, \mathbb{P}(\mathbf{Y}) = -\mathrm{log} \, \mathbb{P}(\mathcal{A} \mathbf{V}),
\end{equation}
where $\mathbf{V} \in \mathbb{R}^{n \times D}$ is the affine transformation of the token sequence $\mathbf{X}$, while $\mathcal{A}$ refers to the attention map for the $\mathit{cls}$ token, calculated as follows: 
\begin{equation} \label{eq:44}
    {
    \mathcal{A} = \mathrm{softmax}(\frac{\mathbf{Q}_{\mathrm{cls}} \, \mathbf{K}^T}{\sqrt{D}}) \in \mathbb{R}^{1 \times n},  \quad
    \sum_{i=1}^{n} \mathcal{A}_{1,i} = 1.
    }
\end{equation}
We let semantic value of $i$-th token as its attention weight $\mathcal{A}_{1,i}$. As we have no access to probability distribution of $\mathbf{V}$, we use a meandering way to approximate the solution. Inspired by the vector quantization~\cite{esser2021taming,van2017neural}, one common sense can be summarized as follows. In the context of neighbourhood, continuous variables can be approximated by discrete quantities, illustrating that if we place the perspective in a local neighborhood, vectors that are closer in distance possess similar semantics. Thus, we can make assumptions of local continuity for $\mathcal{I}$.

\noindent \underline{\textit{Proposition}}. Under the metric space defined by $L_2$ norm and $R^n$, for all $\eta_0 \in R_+$, there exists $\epsilon>0$ for semantic-rich $y_1\in R^n$, such that for all $y_2\in R^n$ satisfying $\Vert y_1 - y_2 \Vert_2 < \epsilon$, there exists $\eta \in R_+ < \eta_0$ to verify the following inequality (see detailed deductions in supplementary materials):
\begin{equation} \label{eq:55}
    {
    \Vert \mathcal{I}(y_1) - \mathcal{I}(y_2) \Vert \ \leq \  \eta. 
    }
\end{equation}

Denote $\mathbf{Y}'$ the $\mathit{cls}$ token after pruning the tokens $\{x_{j, j\in \psi'}, \psi' \subseteq \{1,...,n\}\}$, satisfying that for $\epsilon'$ associated with $\eta_0'<<1$,
\begin{equation} \label{eq:12}
    {
    \Vert \sum_{j \in \psi'} \mathcal{A}_{1,j} \mathbf{V}_j \Vert_2 < \epsilon',
    }
\end{equation}
Then we can deduce from Eq.\,({\ref{eq:55}}) (deduction in supplementary materials) that
\begin{equation} \label{eq:13}
    {
    \mathcal{I}(\mathbf{Y}) \approx \mathcal{I}(\sum_{i=1}^n \mathcal{A}_{1,i} \mathbf{V}_i - \sum_{j \in \psi'} \mathcal{A}_{1,j} \mathbf{V}_j) = \mathcal{I}(\mathbf{Y}').
    }
\end{equation}

Such analysis indicates that, pruning/merging tokens with sufficiently small semantic value, will barely affect the informativity of the $\mathit{cls}$ token.

Semantic value captures token importance/relevance to the overall semantics. We inspect it into data de-redundancy, by giving tokens with high semantics less weights. This ensures tokens with high semantic contributions are more likely to be retained, even if their mutual redundancy is high. In Figure~\textcolor{red}{4} and Section~\ref{sec:abla}, we prove the properties of semantic value, and the existence of subset $\psi'$.

\noindent \underline{\textit{Remark.}} For uni-modal tasks, we adopt intra-modal \textit{cls} token as guidance; while for multi-modal tasks, we use intra-modal or cross-modal \textit{cls} token for semantic value guidance depending on architectures (self-/cross-attention). With such designs, we can leverage cross-modal information for data de-redundancy.

\subsection{Empirical Evaluation of Informativity}  
Before, we identify mutual redundancy and semantic value as key factors for data de-redundancy. Here, we quantitatively measure informativity in the output token sequence $\mathbf{X}=[x_1,x_2,...,x_n]$ of one attention layer.

On the one hand, for mutual redundancy $\mathcal{R}$, we define token redundancy to measure information similarity between tokens: $\widehat{\mathcal{R}} = \mathrm{Avg}\{\mathcal{S}(x_i, x_j)\}$, where $\mathrm{Avg}$ is the averaging operation. On the other hand, to verify the long-tail distribution of semantic value $\mathcal{A}$, we define attention concentration $\widehat{\mathcal{A}}$ as the semantic proportion in the top $1/4$ tokens with highest values: $\widehat{\mathcal{A}}=\sum_{i=1}^{\lfloor n/4 \rfloor}\mathcal{A}_{1,i}$. A larger $\widehat{\mathcal{A}}$ implies a higher degree of concentration on few tokens, ensuring safe reduction to insignificant trailing tokens without losing much information.

To view underlying redundancies in VLMs, we do experiments on BLIP~\cite{li2022blip}, using the parameters fine-tuned on COCO. Figure~\ref{fig:evaluation} counts token redundancy and attention concentration. We empirically draw the conclusions:

\noindent $\bullet$ \textbf{Basis}: token redundancy and attention concentration are consistently high in most blocks: $95\%$ in intermediate layers, the left $3/4$ tokens contributes only $5\%$.

\noindent $\bullet$ \textbf{Trend}: token redundancy gradually increases as blocks deepen, implying a tendency of clustering done by the network. Attention concentration presents a bottleneck trend, with the maximum value reached at the intermediate layer. 

These observations support the fact that tokens in the attention layer possess high mutual redundancy and concentrated semantic value, thus acceleration by data de-redundancy is promising. Next, we achieve the goal by one novel Turbo.

\subsection{Plug-and-Play Turbo} 
\noindent {\bf Strategy.} As mutual redundancy $\mathcal{R}$ and semantic value $\mathcal{A}$ are two key factors to keep information and remove redundancy, we propose information degree $\mathcal{E}$ for balancing, by two fusion strategies: weighted difference or coupled division:
\begin{equation}  \label{eq:balance}
    {\mathcal{E} = \mathcal{R}-\alpha\mathcal{A}, \quad \mathcal{E} = \mathcal{R}/{\mathcal{A}}.
    }
\end{equation}
Table~\ref{tab:degree} compares in detail to show the significance of trade-offs between $\mathcal{R}$ and $\mathcal{A}$. With information degree $\mathcal{E}$, we naturally propose Turbo to accelerate by data de-redundancy. Since the data organization of most VLMs is the attention-based sequence, Turbo could perform on any block with great flexibility.

\noindent {\bf Role.} Turbo behaves in differentiated forms for generation and understanding. For high-level understanding (classification, retrieval, caption, VQA), Turbo calculates information degree for $n$ input tokens $\mathbf{X} \in \mathbb{R}^{n \times D}$ in each layer after the attention block, then sorts them to merge the top-level ones into the rest $\mathbf{X}{'}$.  
\begin{equation}  \label{eq:highlevel}
    {\mathbf{X}{'} =\Phi_{\mathrm{turbo}}^{\mathrm{high}}(\mathbf{X}, \; \Upsilon) = \Psi_{\mathrm{meg}}(\Psi_{\mathrm{sort}}(\mathcal{E}(\mathbf{X})), \; \Upsilon)
    \in \mathbb{R}^{(n-\Upsilon) \times D}, 
    }
\end{equation} 
where $\Psi_{\mathrm{sort}}$ refers to the sort operation based on the information degree, $\Psi_{\mathrm{meg}}$ is the merging operation, and $\Upsilon$ is the drop ratio to all sequential tokens. For parallel computing, we set $\Upsilon$ to one constant in each batch.

For generative tasks (text-to-image and image-to-image), fine-grained pixel requirements make such simple merging of redundant tokens infeasible. We hence divide the Turbo into more modules, making its role become \underline{merge \& restore}. Specifically, before each VLMs' block, Turbo calculates and records information degree between all tokens, then merges redundant tokens using $\Psi_{\mathrm{meg}}$; while after each VLMs' block, Turbo restores the merged tokens by $\Psi_{\mathrm{rest}}$, that is, weighted sum related and unmerged tokens, with reference to information degree. 
\begin{equation}  \label{eq:lowlevwl}
    {\mathbf{X}{'} = \Phi_{\mathrm{turbo}}^{\mathrm{low}}(\mathbf{X}, \; \Upsilon) = 
    \Psi_{\mathrm{rest}}(\Psi_{\mathrm{meg}}(\Psi_{\mathrm{sort}}(\mathcal{E}(\mathbf{X})), \Upsilon), \; \mathcal{E}(\mathbf{X})) \in \mathbb{R}^{n \times D}. 
    }
\end{equation}
Such Turbo greatly reduces computing overhead for VLMs' blocks, while ensuring pixel-level generation. Besides that, Turbo also allows the bulk of computing to be done in parallel, making it friendly to the modern GPU devices. For more implementation details, please refer to supplementary materials.

\noindent \textbf{Remark.} Turbo wins many pros, compared to existing acceleration. \underline{\textit{Universality}}. It performs data de-redundancy, {\em i.e.}, reducing computing from source of the AI system. And the resultant data could be generally used for various VLMs, {\em e.g.}, understanding/generation, multi-/uni-modality, showing powerful compatibility. \underline{\textit{Practicality}}. It fully considers the potential data-repetition guided by semantics, accelerating with little performance loss. Besides, it's user-friend without cumbersome tricks. \underline{\textit{Plug-and-Play}}. It works as plug-in without additional parameters for re-training, and is plain to avoid trivial development efforts.

\begin{table}[t]
\scriptsize
\caption{\textbf{NLVR Acceleration for BLIP~\cite{li2022blip} on NLVR2 dataset.} Ratio $\Upsilon$ denotes the number of reduced tokens per layer. We also apply Turbo to the VLMs pruned by UPop, proving that our Turbo is perpendicular to the model-perspective accelerations. Turbo surpasses other methods by a large margin on inference speed and accuracy.}
\centering
\begin{tabular}{C{2.2cm}|C{1cm}C{1.1cm}|C{1cm}|C{2cm}C{2cm}}
\toprule
\multirow{2}{*}{Plug-In} & \multicolumn{2}{c|}{Performance} & \multirow{2}{*}{\begin{tabular}[c]{@{}c@{}}Ratio\\ $\Upsilon$ \end{tabular}} & \multicolumn{2}{c}{Acceleration} \\ \cline{2-3} \cline{5-6} 
                        & Dev        & Test        &                        & FLOPs       & Throughput         \\ \hline \hline
-                       & 82.5          & 83.3           & -                      & 132.5       & 117.5              \\ \hline
UPop~\cite{shi2023upop}                   & 80.3          & 81.1            & -                      & 89.4 (-0.32×)       & 138.5 (1.18×)      \\
Redundancy                    & 79.3          & 79.5           & 35                     & 77.2 (-0.42×)        & 182.9 (1.56×)      \\
Semantics                    & 80.2         & 81.0           & 35                     & 77.2 (-0.42×)        & 186.7 (1.59×)      \\ \hline \hline
\multirow{2}{*}{Turbo}  & \textbf{81.4}          & \textbf{82.2}           & 35                     & 77.2 (-0.42×)       & 182.8 (1.56×)      \\
                        & 80.5           & 81.5           & 45                     & 62.2 (-0.53×)        & \textbf{224.2 (1.91×)} 
                        \\ \hline 
Turbo+UPop                       & 79.0          & 79.8           & 35           & \textbf{54.7 (-0.59×)}      & 188.2 (1.60×)
                        \\ \bottomrule
\end{tabular}
\label{tab:nlvr}
\end{table}

\begin{table}[t]
\scriptsize
\caption{\textbf{VQA Acceleration for BLIP~\cite{li2022blip} on VQAv2 dataset.} Through applying $\Upsilon=60$, we achieve the best trade-off between performance reservation and acceleration.}
\centering
\begin{tabular}{C{2.2cm}|C{1cm}C{1.1cm}|C{1cm}|C{2cm}C{2cm}}
\toprule
\multirow{2}{*}{Plug-In} & \multicolumn{2}{c|}{Performance} & \multirow{2}{*}{\begin{tabular}[c]{@{}c@{}}Ratio\\ $\Upsilon$ \end{tabular}} & \multicolumn{2}{c}{Acceleration} \\ \cline{2-3} \cline{5-6} 
                        & Dev        & Std       &                        & FLOPs           & Throughput     \\ \hline \hline
-                       & 77.4            & 77.5           & -                      & 92.1            & 148.1          \\ \hline
UPop~\cite{shi2023upop}                    & 76.3            & 76.3           & -                      & 65.2 (-0.30×)   & 167.8 (1.13×)  \\
Redundancy                    & 74.8            & 74.8           & 40                     & 65.7 (-0.29×)   & 184.8 (1.25×)  \\
Semantics                    & 75.3            & 75.3           & 40                     & 65.7 (-0.29×)   & 188.2 (1.27×)  \\ \hline \hline
\multirow{2}{*}{Turbo}  & \textbf{76.8}            & \textbf{76.9}           & 40                     & 65.7 (-0.29×)   & 184.6 (1.25×)  \\
                        & 76.6            & 76.7           & 60                     & 52.5 (-0.43×)   & \textbf{232.7 (1.57×)}  
                                                \\ \hline 
Turbo+UPop               & 75.4          & 75.4           & 40          & \textbf{44.8 (-0.51×)}      & 206.8 (1.40×)
                        \\ \bottomrule
\end{tabular}
\label{tab:vqa}
\end{table}

\section{Experiments}
\noindent \textbf{Understanding Benchmarks.} We evaluate Turbo on uni-modal Image Classification, and multi-modal tasks like Cross-modal Retrieval, Natural Language for Visual Reasoning, Visual Question Answering and Image Captioning. We experiment image classification with AugReg~\cite{steiner2021train} and SWAG~\cite{singh2022revisiting} on Imagenet-1k; multi-modal tasks with BLIP~\cite{li2022blip}/BLIP2~\cite{li2023blip} on Flickr30k, COCO and NLVR2 and MiniGPTv2~\cite{chen2023minigpt} on Vizwiz~\cite{bigham2010vizwiz}, OKVQA~\cite{marino2019ok} and GQA~\cite{hudson2019gqa}.

\noindent \textbf{Generation Benchmarks.} We evaluate on Stable Diffusion 1.5~\cite{rombach2022high} by generating $2000$ images, each resolution is set to $512*512$. The text classes used are from ImageNet-1k. All experiments are conducted on one 3090 GPU. For more details on different benchmarks, please refer to supplementary materials.

\subsection{Comparison with State-Of-The-Art} 
We carry out full experiments on uni-/multi-modal VLMs, to prove effectiveness. Two typical methods in the model-perspective acceleration, {\em i.e.}, GPTQ~\cite{frantar2022gptq} for quantization and UPop~\cite{shi2023upop} for model pruning, are used for full comparisons. In addition, we introduce two baselines: Redundancy and Semantics, to represent methods with only mutual redundancy and semantic value.

\begin{table}[t]
\scriptsize
\caption{\textbf{Caption Acceleration on COCO dataset.} The drop ratio is 16 and (30, 40) for Turbo on BLIP2 and BLIP respectively. Turbo attains competitive performance while largely enhance throughput. We also leverage Turbo upon GPTQ (4-bit model quantization) and UPop (model pruning) for further inference acceleration.}
\setlength\tabcolsep{4pt}
\centering
\begin{tabular}{C{1.3cm}|C{2.2cm}|C{1cm}C{1.3cm}|cc}
\toprule
\multirow{2}{*}{VLMs} & \multirow{2}{*}{Plug-In} & \multicolumn{2}{c|}{Performance} &  \multicolumn{2}{c}{Acceleration} \\ \cline{3-4} \cline{5-6} 
          &               & B@4        & CIDEr                & FLOPs       & Throughput      \\   
\hline \hline
\multirow{6}{*}{BLIP~\cite{li2022blip}}  & -      & 39.7 & 133.3 & 330.2  & 34.2         \\
                       & UPop~\cite{shi2023upop}   & \textbf{38.6} & 128.9 & 137.9 (-0.58×)  & 56.6 (1.65×) \\
                       & Redundancy   & 35.5 & 120.9 & 134.2 (-0.60×)  & 67.9 (1.99×) \\
                       & Semantics   & 36.4 & 123.8 & 134.2 (-0.60×)  & 70.3 (2.06×) \\  \cline{2-6}
                       & Turbo  & 38.2 & \textbf{130.0} & 134.2 (-0.60×)  & 67.6 (1.98×) \\
                       & Turbo+UPop & 37.3 & 126.0 & \textbf{60.8 (-0.82×)}  & \textbf{74.8 (2.19×)} \\ \hline \hline
\multirow{6}{*}{BLIP2~\cite{li2023blip}} & -      & 42.7 & 145.7 & 1379.2 & 29.5         \\
                       & GPTQ~\cite{frantar2022gptq} & 42.2 & 144.8 & 1379.2 & 30.1 (1.02×) \\
                       & Redundancy   & 40.8 & 137.7 & 1029.5 (-0.25×) & 51.1 (1.73×) \\
                       & Semantics   & 40.3 & 137.9 & 1029.5 (-0.25×) & 52.2 (1.77×) \\  \cline{2-6}
                       & Turbo  & \textbf{42.1} & \textbf{142.2} & 989.7 (-0.29×)  & 50.9 (1.72×) \\ 
                       & Turbo+GPTQ  & 41.4 & 142.0 & \textbf{989.7 (-0.29×)}  & \textbf{52.5 (1.78×)} \\ 
\bottomrule
\end{tabular}
\label{tab:caption}
\end{table}

\begin{table}[t]
\scriptsize
\caption{\textbf{Acceleration on MiniGPTv2~\cite{chen2023minigpt}.} We utilize LLaMA2-chat 7B as the LLM backbone, and 8-bit quantization for GPTQ~\cite{frantar2022gptq}, and evaluate on the VQA benchmarks (accuracy as evaluation metric). Turbo still exhibits superior performance, and it can be orthogonal to further boost the model-perspective acceleration.} 
\setlength\tabcolsep{4pt}
\centering
\begin{tabular}{c|ccc|c|cc}
\toprule
\multirow{2}{*}{Plug-In} & \multicolumn{3}{c|}{Dataset} & \multirow{2}{*}{\begin{tabular}[c]{@{}c@{}}Ratio\\ $\Upsilon$ \end{tabular}} & \multicolumn{2}{c}{Acceleration} \\ \cline{2-4} \cline{6-7} 
               & OKVQA        & GQA      &   Vizwiz  &      & FLOPs       & Throughput      \\   
\hline \hline
 -      & 57.9 & 59.9 & 56.9 &  - &  5736.5 & 6.8         \\
GPTQ~\cite{frantar2022gptq}   & 57.7 & 59.6 & 55.9 & -  & 5736.5  &  5.9 (0.87×) \\
Redundancy   & 54.5 & 52.8 & 55.1 &  12 &  4411.3 & 9.7 (1.43×) \\
Semantics   & 54.3 & 53.2 & 55.2 & 12 & 4411.3  & 9.8 (1.44×) \\ \hline \hline
\multirow{2}{*}{Turbo}  & \textbf{56.3} & \textbf{56.7} & \textbf{55.7} &  12 & 4411.3  & 9.7 (1.43×) \\
 & 54.7 & 54.2 & 55.2 &  16 &  \textbf{3853.4} & \textbf{11.8 (1.74×)} \\ \hline
Turbo+GPTQ & 56.1 & 56.6 &  55.7 & 12 & 4411.3  &  8.2 (1.21×)\\
\bottomrule
\end{tabular}
\label{tab:minigpt}
\end{table}

\noindent \textbf{Multi-Modal VLMs for Understanding.} Thorough experiments on a wide range of multi-modal tasks for VLMs BLIP~\cite{li2022blip}, BLIP2~\cite{li2023blip} and MiniGPTv2~\cite{chen2023minigpt} are done to test the effectiveness, in Tables~\ref{tab:nlvr},~\ref{tab:vqa},~\ref{tab:caption},~\ref{tab:minigpt} and~\ref{tab:retrieval}. On all benchmarks, Turbo achieves the best trade-off for acceleration and performance preservation. We achieve better or competitive performance with less time, than the quantization (GPTQ) and the model pruning idea (UPop). Compared to model-perspective accelerations that require re-training for specific VLMs, Turbo is training-free to serve as universal plug-in for various VLMs. What's more, due to the huge cost of memory access time, FLOPs is not inversely proportional to actual throughput. This leads to poor accelerations for model-perspective methods in reality, since these methods can't reduce the memory access time of data. For quantization-based method, due to the dequantization process, throughput could even be lower than the original one. On the contrary, by directly merging data, our Turbo saves both memory access time and calculation amount.

\noindent \underline{NLVR}: Table~\ref{tab:nlvr} reports results (accuracy) under different drop ratios $\Upsilon$. Turbo performs the best even with an acceleration of $1.91$× compared to $1.18$× for UPop. We also insert Turbo into VLMs pruned by UPop, showing that Turbo is perpendicular to model-perspective acceleration, reflecting our universality.

\noindent \underline{VQA}: Table~\ref{tab:vqa} \& Table~\ref{tab:minigpt} evaluate the VQA task (accuracy) for BLIP~\cite{li2022blip} and LLM-based MiniGPTv2~\cite{chen2023minigpt}. In general, Turbo again exceeds the other methods either on performance or acceleration, even with several large drop ratios.

\begin{table*}[t]
\scriptsize
\caption{\textbf{Retrieval Acceleration on Flickr30K/COCO datasets.} The drop ratio is 16 and (30,40) for Turbo on BLIP2 \& BLIP. Turbo gets superior results, with a drop of $0.4\%$ on BLIP image-text retrieval, while other methods drops $3.5\%$, $4.0\%$ and $4.7\%$.}
\centering
\begin{tabular}{C{1.6cm}|C{1.4cm}|C{1.6cm}|C{0.9cm}C{0.8cm}|C{0.9cm}C{0.8cm}|C{1.2cm}C{1.8cm}}
\toprule
\multirow{2}{*}{VLMs} & \multirow{2}{*}{Dataset}   & \multirow{2}{*}{Plug-In} & \multicolumn{2}{c|}{Image-to-Text} & \multicolumn{2}{c|}{Text-to-Image} & \multicolumn{2}{c}{Acceleration} \\ \cline{4-9}
                       &                            &                         & R@1            & R@5            & R@1            & R@5            & FLOPs           & Throughput     \\ \hline \hline 
\multirow{8}{*}{BLIP2~\cite{li2023blip}} & \multirow{4}{*}{Flickr30k} & -                       & 97.6           & 100.0          & 89.7           & 98.1           & 717.5                  & 10.7                        \\ \cline{3-9} 
                       &                            & Redundancy                    & 96.4           & 100.0          & 86.5           & 97.2           & 376.0        & 19.9               \\  
                       &                            & Semantics                    & 96.0          & 100.0          & 86.2           & 97.2           & 376.0        & \textbf{20.2}               \\
                       &                            & Turbo                   & \textbf{96.7}    & \textbf{100.0}    & \textbf{87.9}     & \textbf{97.4}     & \textbf{376.0 }         & 19.7               \\ \cline{2-9} 
                       & \multirow{4}{*}{COCO}      & -                       & 85.4           & 97.0           & 68.3           & 87.7           & 717.5                  & 10.8                        \\  \cline{3-9}
                       &                            & Redundancy                    & 83.6           & 95.8           & 66.4           & 86.7           & 396.5         & 18.7               \\
                       &                            & Semantics                    & 83.2           &   95.6         &   66.0         &    86.3        & 396.5         & \textbf{19.0 }               \\
                       &                            & Turbo                   & \textbf{84.2}     & \textbf{96.2}     & \textbf{67.1}     & \textbf{87.1}     & \textbf{396.5}         & 18.6                \\ \hline \hline
\multirow{12}{*}{BLIP~\cite{li2022blip}} & \multirow{6}{*}{Flickr30k} & -                       & 97.2           & 99.9           & 87.3           & 97.6           & 55.5                   & 281.0                       \\ \cline{3-9} 
                       &                            & Redundancy                    & 93.7           & 99.5           & 80.1           & 95.5           & 37.0          & 375.6              \\ 
                       &   & Semantics                    & 92.8           & 99.4           & 79.3           & 95.0           & 37.0          & 380.0              \\ 
                       &                            & UPop~\cite{shi2023upop}                    & 92.5           & 99.0           & 78.4           & 94.5           & 39.1          & 322.2               \\  \cline{3-9} 
                       &                            & \multirow{2}{*}{Turbo}                  & \textbf{96.8}          & \textbf{99.8}     & \textbf{85.1}           & \textbf{96.8}          & 37.0           & 375.0               \\
                       &                            &                    & 94.2           & 99.4           & 82.9           & 96.0           & \textbf{31.1 }          & \textbf{449.2 }              \\ \cline{2-9} 
                       & \multirow{6}{*}{COCO}      & -                       & 81.9           & 95.4           & 64.3           & 86.1           & 55.5                   & 285.0                       \\ \cline{3-9} 
                       &                            & Redundancy                    & 74.2           & 92.1           & 56.5           & 80.4           & 36.8           & 381.3               \\ 
                       &   & Semantics                    & 75.3           & 92.6           & 57.2           & 80.9           & 36.8           & 385.8               \\
                       &                            & UPop~\cite{shi2023upop}                    & 77.4           & 93.4           & 59.8           & 83.1           & 39.1          & 323.2              \\ \cline{3-9} 
                       &                            & \multirow{2}{*}{Turbo}                   & \textbf{78.8}           & \textbf{94.7}           & \textbf{61.3}           & \textbf{84.2}          & 36.8         & 380.5              \\
                       &                            &                    & 77.5           & 93.8           & 60.5           & 83.7           & \textbf{34.1}          & \textbf{424.3}
                       \\ \bottomrule
\end{tabular}
\label{tab:retrieval}
\end{table*}

\noindent \underline{Image Captioning} requires fine-grained information. Table~\ref{tab:caption} reports results on BLIP and BLIP2. Relying solely on mutual redundancy or semantic value for token merging performs poorly, implying non-negligible information loss. In contrast, Turbo retains much better results with same FLOPs and throughput, proving that it keeps key information by merging background rather than foreground. Besides, Turbo could further accompany these model-perspective accelerations (such as UPop \& GPTQ) to get better efficiency.

\noindent \underline{Multi-Modal Retrieval} often deals with large amount of data, {\em i.e.} millions or even billions of image-text pairs in real scenarios, so there is an urgent demand for inference speed-up. Table~\ref{tab:retrieval} demonstrates that, Turbo exceeds model-perspective methods and baselines by a large margin, {\em i.e.} $4.3\%$ for image-text and $6.7\%$ for text-image on BLIP Flickr30k compared to UPop~\cite{shi2023upop}. Such acceptable accuracy drops show our method helps to make large-scale retrieval efficiently.

\begin{table}[t]
\scriptsize
\caption{\textbf{Generation Acceleration for Stable Diffusion.} Turbo wins higher quality (FID) and faster throughput, regardless of text-to-image or image-to-image tasks.}
\centering
\begin{tabular}{C{1.8cm}|C{1.2cm}|C{1.2cm}C{2cm}|C{1.2cm}C{2cm}}
\toprule
\multirow{2}{*}{Plug-In} & \multirow{2}{*}{Ratio $\Upsilon$} & \multicolumn{2}{c|}{Text-to-Image} & \multicolumn{2}{c}{Image-to-Image} \\ \cline{3-6} 
                        &                                                                    & FID           & Throughput         & FID           & Throughput         \\ \hline  \hline
-                       & -                                                                  & 32.12         & 0.32               & 30.04         & 0.29               \\ \hline 
\multirow{2}{*}{Redundancy}   & 10                                                                 & 32.80         & 0.39 (1.22×)              & 30.69         & 0.37 (1.28×)              \\
                        & 20                                                                 & 32.86         & 0.44 (1.38×)              & 30.75         & 0.40 (1.38×)            \\ \hline 
\multirow{2}{*}{Turbo}  & 20                                                                 & \textbf{32.63}         & 0.43  (1.34×)             & \textbf{30.48}         & 0.40 (1.38×)              \\
                        & 30                                                                 & 32.77         & \textbf{0.50 (1.56×)}             & 30.56         & \textbf{0.49 (1.69×)}              \\ \bottomrule
\end{tabular}
\label{tab:sdgene}
\end{table}

\noindent \textbf{Multi-Modal Generation.} Table~\ref{tab:sdgene} compares across text-to-image, image-to-image tasks, for fine-grained visual generation. Comparing to acceleration only by mutual redundancy, Turbo achieves better generation quality, {\em i.e.}, lower FID scores. While getting similar generation results, Turbo brings great throughput gains over vanilla stable diffusion models.

\noindent \textbf{Uni-Modal Foundation Models.} On ImageNet-1k, we experiment with two uni-modal training plans, {\em i.e.}, AugReg~\cite{steiner2021train} and SWAG~\cite{singh2022revisiting}. Table~\ref{tab:uni} compares Turbo with one prevalent baseline: ToMe~\cite{bolya2022token}. Note that image classification is rather simple comparing to multi-modal understanding tasks, so the performance drop is relatively small for the baseline. Nevertheless, Turbo surpasses ToMe on models of all sizes and training plans, while keeping the same acceleration rate.

\begin{table}[t]
\scriptsize
\caption{\textbf{Uni-Modal Visual Classification on ImageNet.} Comparing to ToMe~\cite{bolya2022token}, Turbo shows better results on all-size models, while providing same acceleration.} 
\centering
\begin{tabular}{C{1.8cm}|C{2cm}|C{1.3cm}|C{1.7cm}|C{2.3cm}}
\toprule
Model                 & Backbone                  & Plug-In  & Top-1 Acc  & Acceleration           \\ \hline \hline
\multirow{6}{*}{ViT~\cite{steiner2021train}}  & \multirow{2}{*}{ViT-S/16} & ToMe     & 81.4/79.3 & \multirow{2}{*}{1.53×} \\                             \cline{3-4}
                      &                           & Turbo    & 81.4/\textbf{79.9} &                        \\ \cline{2-5} 
                      & \multirow{2}{*}{ViT-B/16} & ToMe     & 84.5/82.6 & \multirow{2}{*}{1.62×} \\ \cline{3-4}
                      &                           & Turbo    & 84.5/\textbf{83.1} &                        \\ \cline{2-5} 
                      & \multirow{2}{*}{ViT-L/16} & ToMe     & 85.8/84.2 & \multirow{2}{*}{1.71×} \\ \cline{3-4}
                      &                           & Turbo    & 85.8/\textbf{84.6} &                        \\ \hline \hline
\multirow{6}{*}{SWAG~\cite{singh2022revisiting}} & \multirow{2}{*}{ViT-B/16} & ToMe     & 85.3/84.5 & \multirow{2}{*}{1.85×} \\ \cline{3-4}
                      &                           & Turbo    & 85.3/\textbf{84.9} &                        \\ \cline{2-5} 
                      & \multirow{2}{*}{ViT-L/16} & ToMe     & 88.1/87.7 & \multirow{2}{*}{1.98×} \\ \cline{3-4}
                      &                           & Turbo    & 88.1/\textbf{87.9} &                        \\ \cline{2-5} 
                      & \multirow{2}{*}{ViT-H/14} & ToMe     & 88.6/88.2 & \multirow{2}{*}{1.91×} \\ \cline{3-4}
                      &                           & Turbo    & 88.6/\textbf{88.4} &                        \\ 
\bottomrule
\end{tabular}
\label{tab:uni}
\end{table}

\subsection{Ablation Study \& Discussion} \label{sec:abla}
We here make comprehensive ablations to dissect components. Without loss of generality, all analysis below are conducted on BLIP image captioning~\cite{li2022blip}.

\noindent \textbf{Effectiveness of Mutual Redundancy \& Semantic Value} are studied in Table~\ref{tab:degree}. As reported in Section~\ref{sec:theory}, mutual redundancy reveals the hidden dependency relationship between tokens. Adding mutual redundancy solely or on semantic value, witnesses obvious boosts in the performance. Moreover, semantic value also has an apparent enhancement for model performance. By balancing two components, we launch the information degree to achieve the best results.

\noindent \textbf{Key Components with Respect to Drop Ratio.} Figure~\textcolor{red}{4} also investigates the performance fluctuation within a wide range of drop ratio $\Upsilon$. (a) Scheme with only mutual redundancy is relatively stable on the large scope but performs badly on small $\Upsilon$. This accords with our prediction in Section~\ref{sec:theory} that mutual redundancy can't distinct important tokens. Merging semantic-rich tokens at early stages inevitably loses fine-grained information. (b) Scheme solely guided by semantic value possesses satisfying results on small $\Upsilon$ but drops dramatically on large $\Upsilon$. This is because that the assumption set in Eq.\,(\ref{eq:12}) no longer holds, resulting in a dramatic drop in performance. By combing semantic value with mutual redundancy, we achieve superior performance on all the scope of $\Upsilon$.

\begin{table}[t]
\scriptsize
\caption{\textbf{Ablation Study on Key Components.} We validate the effectiveness of each component, and compare fusion strategies. This result proves the necessity of both mutual redundancy and semantic value, with a prominent performance boost.}
\setlength\tabcolsep{3.5pt}
\centering
\begin{tabular}{C{0.6cm}|C{2cm}C{1.8cm}C{1.2cm}|C{1cm}cC{1.8cm}}
\toprule 
& Redundancy $\mathcal{R}$ & Semantics $\mathcal{A}$ & Fusion $\mathcal{E}$ & B@4 & CIDEr & Throughput \\ 
\hline \hline
A1 &   &  & - & 34.5 & 112.8 &  73.6 \\
A2 & \checkmark  &  & - & 35.5 & 120.9 & 67.9 \\
A3 &  & \checkmark & - & 36.4 & 123.8 & 70.3 \\
A4 & \checkmark & \checkmark  & $+$ & 38.2 & 130.0 & 67.6 \\
A5 & \checkmark & \checkmark  & $\times$ & 38.2 & 129.9 & 62.8 \\
\bottomrule
\end{tabular}
\label{tab:degree}
\end{table}

\begin{figure}[t]
\centering
\parbox{.42\textwidth}{\caption{\textbf{Ablation Study on Drop Ratio $\Upsilon$.} Semantic value retains superior performance when $\Upsilon$ is small, mutual redundancy possesses better stability on the large $\Upsilon$. By combining these two components, Turbo obtains competitive results and stability on the whole scope.}}
\hspace{\fill}
\parbox{.52\textwidth}{
\includegraphics[width=0.5\textwidth] {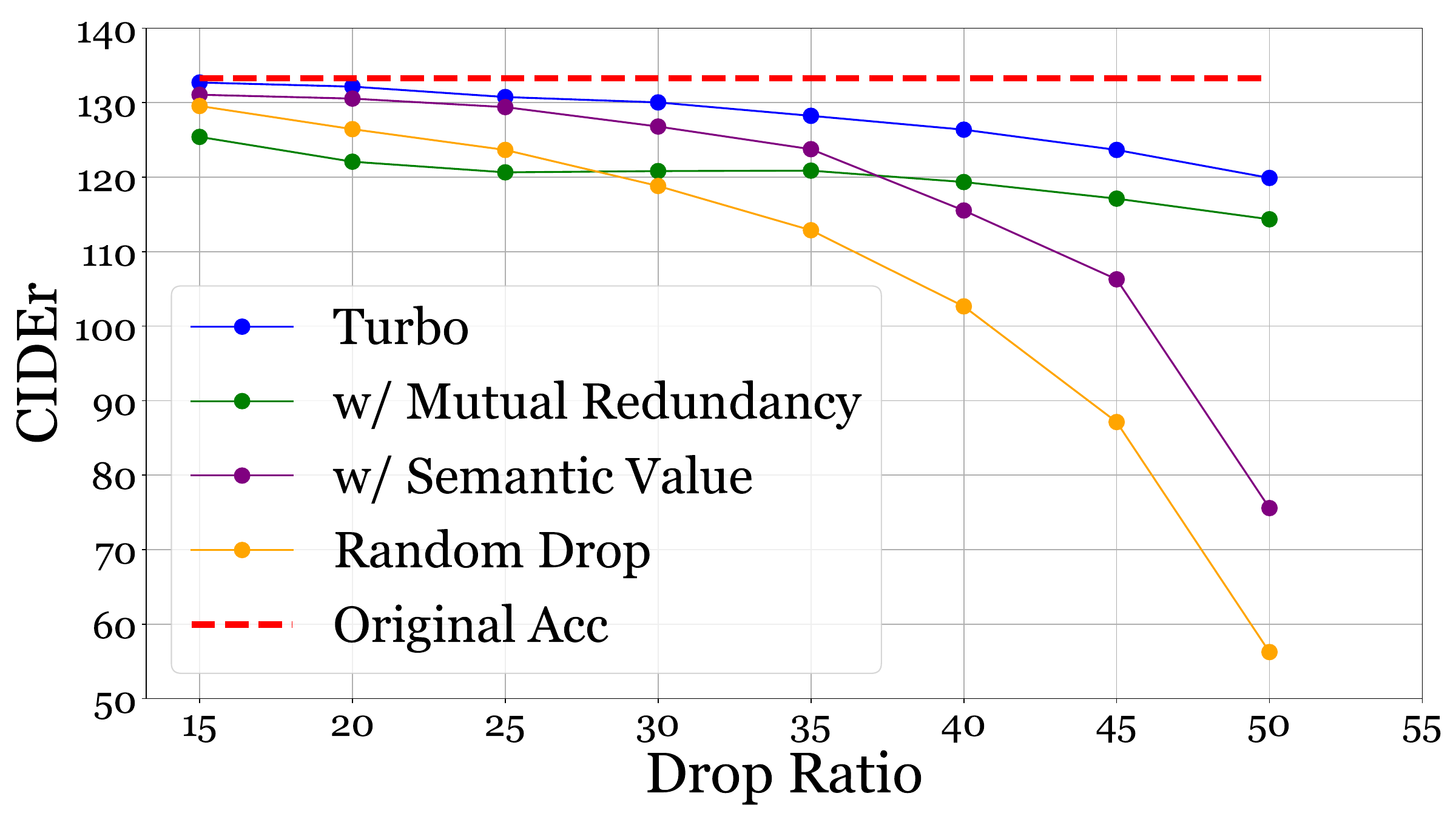}
}
\label{fig:abla}
\end{figure}

\noindent \textbf{Information Degree.} In Eq.\,(\ref{eq:balance}), we adopt two fusion strategies. Table~\ref{tab:degree} shows that coupled division and weighted difference achieve similar performance. Due to the complexity of multiplication over addition, coupled division is slower than weighted difference. Moreover, several complex strategies are studied: dynamic $\alpha$ on different layers (see supplementary materials), but they all result in minor gains, so we use weighted difference for efficient merging.

\noindent \textbf{Robustness.} The balancing coefficient $\alpha$ in Eq.\,(\ref{eq:balance}) may affect the Turbo performance. Hence, Figure~\textcolor{red}{5} experiments to choose $\alpha$ on image captioning for BLIP (VIT-Base and VIT-Large). From the range $\alpha \in \{1,2,...,20\}$, the performance fluctuates slightly on both models with different sizes, indicating that Turbo is unaware of $\alpha$ within a certain range. This proves our robustness.

\begin{figure}[t]
\centering
\parbox{.45\textwidth}{\caption{\textbf{Ablation Study of Balancing Coefficient $\alpha$.} On image captioning using BLIP (VIT-Base and VIT-Large), these results prove our robustness, as the performance varies slightly.}}
\hspace{\fill}
\parbox{.53\textwidth}{
\includegraphics[width=0.5\textwidth] {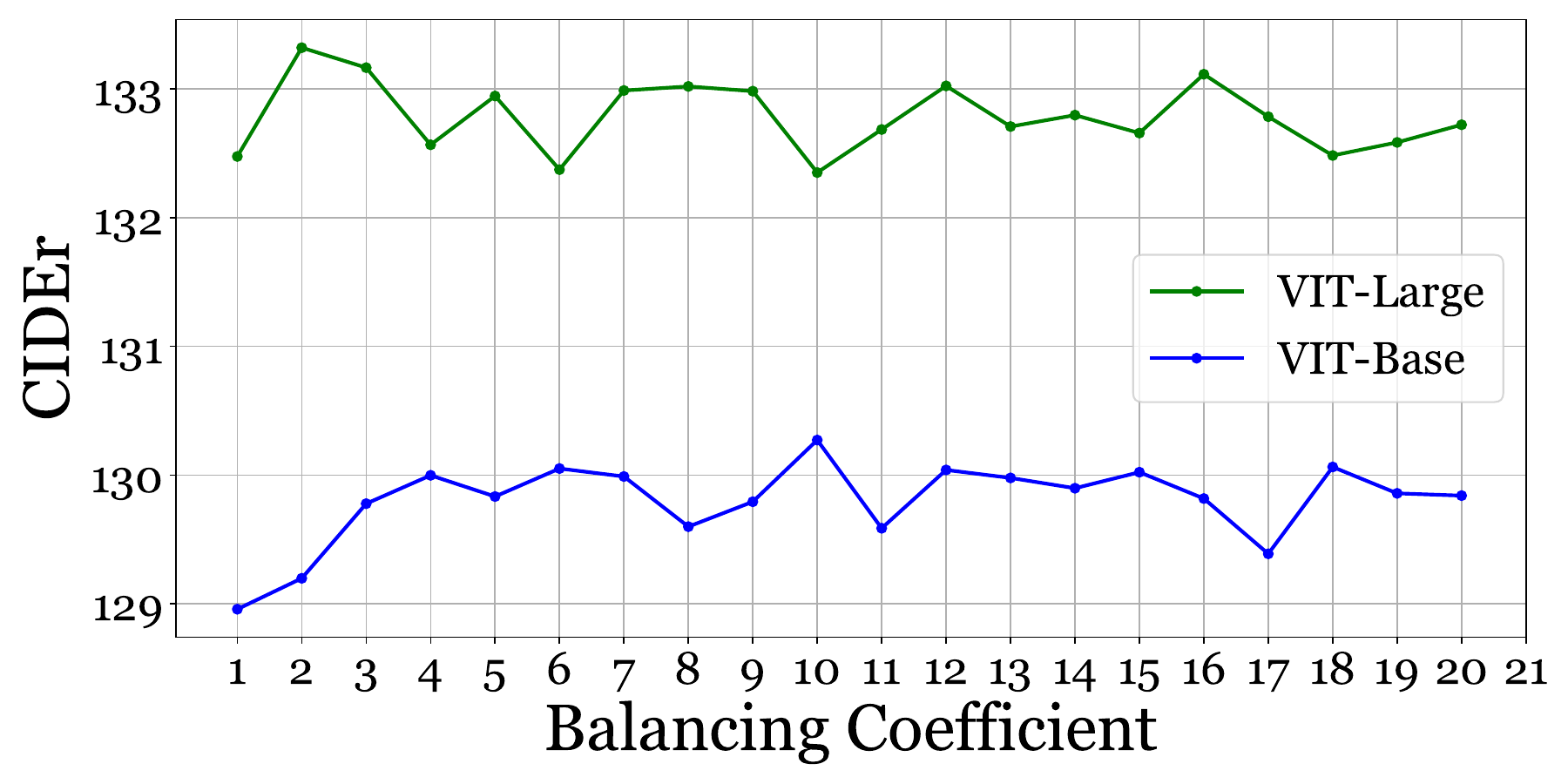}
}
\label{fig:robust}
\end{figure}

\begin{figure}[t]
\begin{center}
\includegraphics[width=0.96\textwidth] {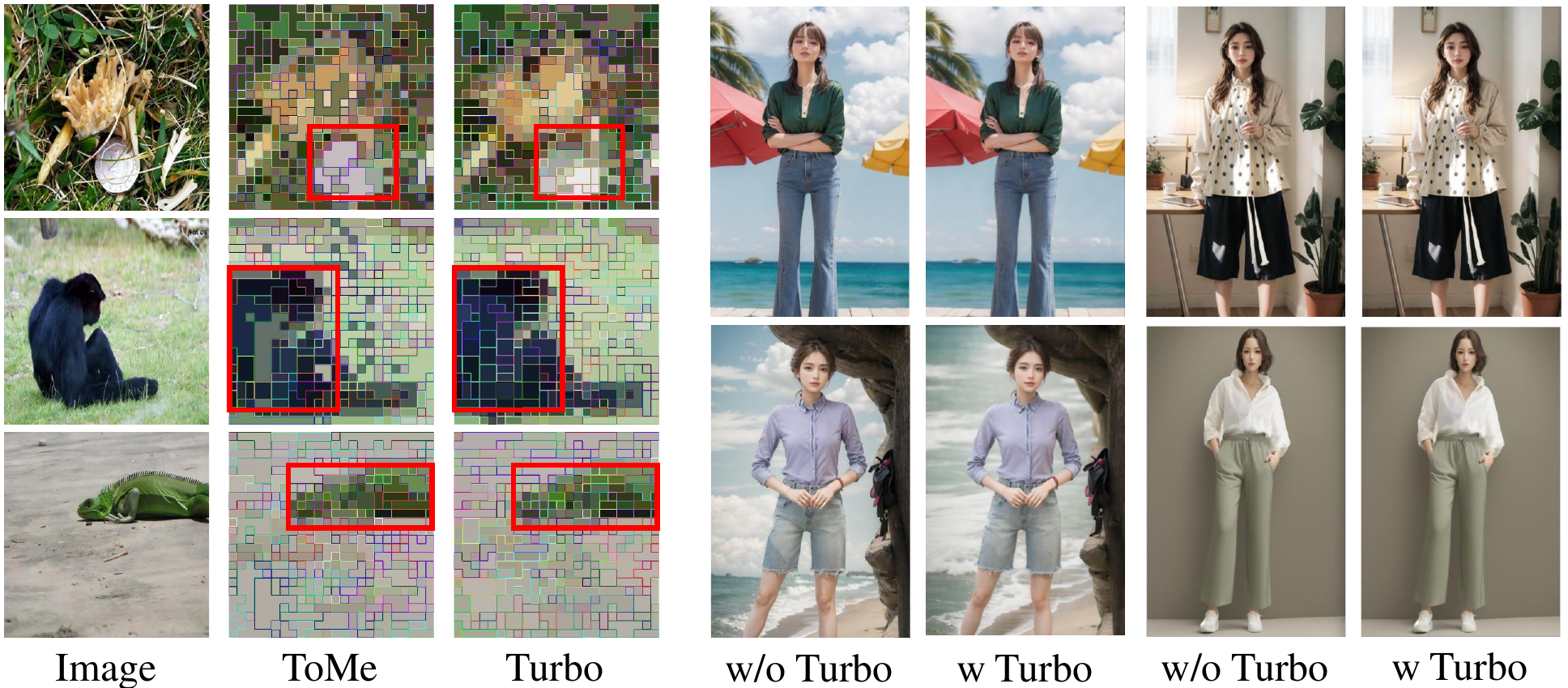}
\end{center}
\caption{\textbf{Visualizations.} \underline{Left}: Turbo merges background patches, while retains foreground patches with semantics, preserving more key information. \underline{Right}: The quality of text-to-image generation is close before and after Turbo acceleration.}
\label{fig:mergepro}
\end{figure}

\subsection{Visualization for Turbo} \label{sec:vis}
\noindent {\bf Image Classification.} Figure~\ref{fig:mergepro} shows the process of token merging, and we highlight foreground in red boxes. Comparing to ToMe~\cite{bolya2022token}, Turbo merges more background patches, while retaining most foreground patches with semantics. Thus, data compressed by Turbo keeps  comprehensive/fine-grained information.

\noindent {\bf Text-to-Image Generation.} \ Figure~\ref{fig:mergepro} shows the results, using text prompts such as: one female model wearing purple long sleeves and blue jeans stands on the coast. Although Stable Diffusion~\cite{rombach2022high} still needs improvement in generation details (such as hands), Turbo acceleration has almost no side effects.

\section{Conclusion}
We propose one novel Turbo plug-in for VLMs' acceleration, from the data perspective. In pursuit of speed-performance trade-offs, Turbo defines information degree for data reduction, taking into account two critical components: mutual redundancy and semantic value. The former evaluates data overlap between sequential tokens; the latter evaluates token's contribution to overall semantics. By eliminating data redundancy from the source, Turbo is generally compatible to various VLMs with trivial development efforts. Extensive experiments show the significance, across multi-modality/uni-modality and understanding/generation.

\noindent \textbf{Acknowledgements.} This work is supported by Alibaba Group through Alibaba Research Intern Program.

\clearpage

In the supplementary material, we first provide more details or ablation studies; then offer theoretical demonstration of the proposed proposition; and finally, we demonstrate some visualization results to justify our superiority.

\section{Implementation Details}
\subsection{Turbo Architecture for Understanding Tasks}
Turbo could be easily plugged in almost any pre-trained, attention-based VLMs to reduce the total sequence length block by block, with no need for further training or adjustment. In practice, we replace all the attention blocks by Turbo. For Turbo, we merge tokens progressively based on their information degree:
\begin{equation} 
    {\mathcal{E} = \mathcal{R}-\alpha\mathcal{A}, \quad \mathcal{E} = \mathcal{R}/{\mathcal{A}}.
    }
\end{equation}

Information degree takes both mutual redundancy and semantic values into account, encouraging insignificant, similar tokens to be merged preferentially. This merging strategy reduces the amount of tokens with duplicated or low informativity, compressing the token sequence with minor information loss. Inspired by~\cite{ahmed2017diverse,bolya2022token}, we follow the bipartite soft matching to calculate mutual redundancy and apply the merging strategy to aggregate tokens. Specifically, due to the over-parameterized problem for token sequence~\cite{marin2021token}, we leverage keys (K) or queries (Q) in QKV attention and the cosine similarity metric between tokens to measure the similarity between tokens. We define mutual redundancy of one token to be the maximum cosine similarity with the other tokens. By adding the quantity of semantic value, which is the attention proportion of each token, the information degree is finally obtained.

\begin{table}[t]
\scriptsize
\setlength\tabcolsep{5pt}
\centering
\caption{\textbf{Ablation Study on Fusion Strategies.} We adopt several fusion strategies and test their performance under different coefficients. Some complex fusion strategies could achieve slightly better results, however, in order to keep the form easy to apply, we adopt the simple weighted average in other experiments.}
\begin{tabular}{c|ccc|ccc}
\toprule 
Strategy & $\alpha$ & $\beta$ & $\gamma$ & B@4 & CIDEr & Throughput \\ 
\hline \hline
S1 & 6 & - & - & 38.2 & 130.0 & 67.6 \\
S2 & -  & - & - & 38.2 & 129.9 & 62.8 \\
S3 & 1 & - & - & 37.8 & 128.5 & 61.9\\
S3 & 2 & -  & - & 37.9 & 128.4 & 61.9 \\
S3 & 3 & -  & - & 36.9 & 125.7 & 61.9 \\
S3 & 4 & -  & - & 36.1 & 123.1 & 61.9 \\
S4 & 6 & 0.9 & 1 & 38.2 & 129.5 & 67.4\\
S4 & 6 & 0.9  & 2 & 38.3 & 129.5 & 67.4\\
S4 & 6 & 0.9  & 3 & 38.1 & 129.5 & 67.4\\
S4 & 6 & 0.9  & 4 & 38.2 & 129.5 & 67.4\\
S4 & 6 & 0.9  & 5 & 38.1 & 129.7 & 67.4\\
S4 & 6 & 0.9 & 6 & 38.0 & 129.2 & 67.4\\
S4 & 6 & 0.9  & 7 & 38.0 & 129.5 & 67.4\\
S4 & 6 & 0.9  & 8 & 38.4 & 130.2 & 67.4\\
S4 & 6 & 0.9  & 9 & 38.2 & 129.9 & 67.4\\
S4 & 6 & 0.9  & 10 & 38.2 & 129.7 & 67.4\\
S4 & 6 & 1.1 & 1 & 38.3 & 130.0 & 67.4\\
S4 & 6 & 1.1  & 2 & 38.3 & 130.3 & 67.4\\
S4 & 6 & 1.1  & 3 & 38.4 & 130.1 & 67.4\\
S4 & 6 & 1.1  & 4 & 38.3 & 129.9 & 67.4\\
S4 & 6 & 1.1  & 5 & 38.5 & 129.9 & 67.4\\
S4 & 6 & 1.1 & 6 & 38.1 & 129.3 & 67.4\\
S4 & 6 & 1.1  & 7 & 38.2 & 129.6 & 67.4\\
S4 & 6 & 1.1  & 8 & 38.5 & 130.3 & 67.4\\
S4 & 6 & 1.1  & 9 & 38.2 & 129.8 & 67.4\\
S4 & 6 & 1.1  & 10 & 38.2 & 129.6 & 67.4\\
\bottomrule
\end{tabular}
\label{tab:strategy}
\end{table}

To avoid excessive computational cost for calculating similarity matrix of the whole token sequence, we leverage bipartite soft matching to speed up the merging process. Suppose the drop ratio is $\Upsilon$, which means we will reduce the amount of tokens by number $\Upsilon$ in each block. In every block, we divide the tokens into two partitions $B$ and $C$ of the same size (if the number of tokens is odd, one partition has 1 more tokens than the other). Then we calculate the mutual redundancy between the two partitions $B$ and $C$. Specifically, for each token in partition $B$, we keep the highest cosine similarity with respect to partition $C$ as its mutual redundancy. After adding the semantic value of each token on partition $B$, we sort the information degree of $B$ and merge the top $\Upsilon$ tokens into $C$, by averaging merging the $\Upsilon$ tokens in $B$ into the corresponding tokens in $C$ with the highest cosine similarity. At last we concatenate the sequence back to continue the forwarding process. In this way, we reduce the length of token sequence by $\Upsilon$ in each block after the attention layer and before the MLP layer. Notice that We call $\Upsilon$ the drop ratio, but it is in fact the number of tokens we reduce every attention block, which is a real ratio by dividing the length of the token sequence. We also note that the semantic value are naturally contained in uni-modal $\mathrm{cls}$ self-attention map or cross-modal $\mathrm{cls}$ cross-attention map depending on the model structure, so we do not need to add additional calculation for the semantic value.

When merging process is finished, some tokens can represent several different patches. This can change the outcome of softmax attention and thus influence the attention calculation. So we fix this with a minor modification:
\begin{equation}
A = \text{softmax} \left( \frac{\mathbf{Q}\mathbf{K}^T}{\sqrt{D}} + \log s \right),
\end{equation}
where $s$ is the number of patches/tokens represented by the merged tokens all along the merging procedure.

In order to avoid excessive merging, {\em i.e.}, merging too many tokens and leave only a few tokens in final blocks. This will cause insufficient expression ability problem and we observe a sharp performance drop at certain drop ratios. So we set up one threshold for the least number of tokens in the final stage, to mitigate the dramatic drop on large $\Upsilon$. Table~\ref{tab:thresh} shows the effectiveness of this restriction for preventing a sudden performance decline.

\subsection{Turbo Architecture for Generation Tasks}
For generative tasks, Stable Diffusion~\cite{rombach2022high} is one popular backbone. Here, Turbo contains one merging module and one inverse-sampling module. For the merging module, we attach Turbo acceleration on the UNet of Stable Diffusion, as UNet consumes the most computation. More specifically, UNet usually consists of three key components: self-attention, cross-attention and FFN. We add Turbo merging/restoring before/after each component. For self-attention and FFN, Turbo merging is calculated by one visual modality; while for cross-attention, Turbo merging is calculated by visual-textual modalities. We evaluate by generating $2000$ images, each resolution is $512*512$. The text classes used are from ImageNet-1k. We use FID scores to metric the generation quality.

\section{More Experiments} 
\subsection{Ablation Study on Fusion Strategy}
We propose four types of fusion strategies to combine mutual redundancy ($\mathcal{R}$) with semantic value ($\mathcal{A}$), as follows:
\begin{align} 
   & \mathrm{S1}: \mathcal{E} = \mathcal{R}-\alpha\mathcal{A}, \\
   & \mathrm{S2}: \mathcal{E} = \mathcal{R}/{\mathcal{A}}, \\
   & \mathrm{S3}: \mathcal{E} = \mathcal{R}(1-\alpha\mathcal{A}), \\
   & \mathrm{S4}: \mathcal{E} = \mathcal{R}-\beta^{\left| \gamma-\mathrm{block\_id}\right|}\alpha\mathcal{A},
\end{align} 
where $\mathrm{S4}$ is designed to allow dynamic scales between $\mathcal{R}$ and $\mathcal{A}$ on different blocks. For example, if $\beta>1$ and $\gamma=6$, then the scale of $\mathcal{A}$ will reach its maximum on block 6 and attain the minimum value on two-end blocks.

As shown in Table~\ref{tab:strategy}, we have done extensive experiments on the four fusion strategies with different coefficients. Though $\mathrm{S4}$ attains the best result, due to its complexity and such slight performance gain, {\em i.e.}, three hyper-parameters to be determined with one gain of only 0.3 on CIDEr, we thus adopt the simple weighted average fusion strategy ($\mathrm{S1}$) on our turbo module.

\begin{table}[t]
\scriptsize
\setlength\tabcolsep{5pt}
\centering
\caption{\textbf{Ablation Study on Threshold.} We validate threshold to prevent dramatic performance drop on large $\Upsilon$. With slight decline on speed, performance with threshold on minimum token length can maintain a smoother decrease when $\Upsilon$ getting larger.}
\begin{tabular}{cc|ccc}
\toprule 
$\Upsilon$ & Threshold & B@4 & CIDEr & Throughput \\ 
\hline \hline
65 & 0 & 28.8 & 95.5 &  107.6 \\
65 & 70 & 31.1 & 104.3 & 103.3 \\
65 & 130 & 34.2 & 115.3 & 100.8 \\
70 & 0 & 27.3 & 89.4 & 113.0 \\
70 & 70 & 31.0 & 103.7 & 110.7 \\
\bottomrule
\end{tabular}
\label{tab:thresh}
\end{table}

\subsection{Ablation Study on Threshold}
When applying large drop ratios $\Upsilon$ on Turbo module, we witness a sharp drop of model performance. We argue that this phenomenon is due to the insufficient expression ability of token sequence length below a certain threshold. So we append a threshold on minimum number of tokens left in the final block. Specifically, we stop the token merging process once the token sequence length is below the threshold. Results in Table~\ref{tab:thresh} demonstrates the effectiveness of such a threshold on large $\Upsilon$. By introducing a threshold to large drop ratio, we improve the model performance by over 10 points on CIDEr with slight acceleration declines.

\subsection{Performance Cap}
To show the upper limit for our method, we also evaluate the performance cap of Turbo under certain performance drop tolerance. Table~\ref{tab:caps} evaluates the Turbo cap with 5\% drops in performance. Turbo can accelerate throughput by up to 2.8 times, almost twice as redundancy-only methods. Such higher caps validate Turbo's power, although the performance sensitivity to acceleration varies among different tasks. Besides, the VLMs' backbones recently continue to grow, causing more data redundancy, so Turbo's cap can be higher in future. 
\begin{table}[t]
\scriptsize
\centering
\caption{\textbf{Acceleration Cap (Performance Drops by 5\%) on BLIP ViT-B.}}
\begin{tabular}{C{3cm}|C{1.2cm}|C{1.2cm}|C{1.5cm}|C{1.8cm}}
\hline 
Method & NLVR & VQA & Caption & Retrieval  \\
\hline
Redundancy $\mathcal{R}$ & 1.5$\times$ & 1.3$\times$ & 1.2$\times$ & 1.1$\times$ \\
\hline
Semantics $\mathcal{A}$ & 1.8$\times$ & 1.4$\times$ & 1.7$\times$ & 1.2$\times$ \\
\hline
\textbf{Turbo} ($\mathcal{R} \, \& \mathcal{A}$)  & \textbf{2.2}$\times$ & \textbf{2.2}$\times$ & \textbf{2.8}$\times$ & \textbf{1.9}$\times$ \\
\hline
\end{tabular}
\label{tab:caps}
\end{table}

Table~\ref{tab:left} also evaluates the Turbo's performance with 25\% tokens left. Turbo accelerates throughput to 1.9 times, with only 2.6\% performance loss (acceptable), on average, which verifies the assumption drawn in main paper that 25\% is sufficient for containing most of the information.

\begin{table}[t]
\scriptsize
\centering
\caption{\textbf{Acceleration of Throughput with 25\% Visual Tokens Left on BLIP.}}
\begin{tabular}{C{1.8cm}|C{2.2cm}|C{2.2cm}|C{2cm}|C{2.2cm}}
\hline 
& NLVR & VQA & Caption & Retrieval \\
\hline
Performance & 81.8/83.3 & 76.9/77.4 & 126.6/133.3 & 94.2/97.2  \\
\hline
Throughput & 212/117 (1.8×) & 185/148 (1.3×) & 96/34 (2.8×) & 449/281 (1.6×) \\
\hline
\end{tabular}
\label{tab:left}
\end{table}

\section{Theoretical Interpretation of the Proposition}
\subsection{Detailed Deduction for Eq.~(9) in The Main Paper}

\noindent \underline{\textbf{Preliminary}}. We first remind the definitions/propositions concerning open ball, neighborhood and continuity in topology.

\noindent \underline{\textit{Definition 1}}. Given a metric space \( (E, d) \), where \(E\) is a set and \(d: E \times E \rightarrow \mathbb{R}\) is a metric on \(E\), the open ball centered at a point \(a \in E\) with radius \(\epsilon > 0\) is defined as the set of all points in \(E\) whose distance to \(a\) is less than \(\epsilon\). Mathematically:
\begin{equation} \label{eq:def}
    B(a, \epsilon) = \{ x \in E : d(a, x) < \epsilon \} 
\end{equation}

\noindent \underline{\textit{Definition 2}}. Let \( (E, d) \) be a metric space, \(a \in E\), we say that \(V\) is a neighborhood of \(a\) in \(E\), and write \(V = \mathcal{V}(a)\) if there exists \(\eta > 0\) such that \(B(a, \eta) \subseteq V\).

\noindent \underline{\textit{Proposition 8}}. Suppose $f : (E, d) \to (E', d')$, then
\begin{equation} \label{eq:prop}
[ f \text{ continue on point } a \in E  ] \iff [ \forall V \in \mathcal{V}(f(a)), \ f^{-1}(V) \in \mathcal{V}(a) ]
\end{equation}

\noindent \underline{\textbf{Demonstration}}. We define $y_1 \in R^n$ as a semantic-rich vector if $y_1$ is in the set of all possible \textit{cls} tokens $\Lambda$. Inspired by the success of quantization methods~\cite{esser2021taming,van2017neural}, $y_1$ can be replaced by the most similar discrete vector in the code-book without losing its informativity. Based on this observation, we make an assumption for the local continuity for informativity $\mathcal{I}$ on $y_1$:

\noindent \underline{\textit{Assumption}}. Under the metric spaces ($R^n$, $L_2$) and ($R$, $L_1$), $\forall y_1 \in \Lambda$, $\mathcal{I} : (R^n, L_2) \to (R, L_1)$ is continue on $y_1$.

According to Proposition~\ref{eq:prop} and Definition~\textcolor{red}{2}, for all $\eta_0 \in R_+$, \(B(\mathcal{I}(y_1), \eta_0) \in \mathcal{V}(\mathcal{I}(y_1))\), so $\mathcal{I}^{-1}(B(\mathcal{I}(y_1), \eta_0)) \in \mathcal{V}(y_1)$, therefore there exists $\epsilon > 0$ such that $B(y_1, \epsilon) \subseteq \mathcal{I}^{-1}(B(\mathcal{I}(y_1), \eta_0))$. By mapping the open ball $B(y_1, \epsilon)$ back using $\mathcal{I}$, we can find $\eta \leq \eta_0$ such that $\mathcal{I}(B(y_1, \epsilon)) \subseteq B(\mathcal{I}(y_1), \eta)$. The operator $\leq$ can be replaced by $<$ by choosing small $\epsilon$, proving the proposition in the main paper.

\subsection{Detailed Deduction for Eq.~(11) in The Main Paper}
Denote $\mathbf{Y}'_0$ the $\mathit{cls}$ token after pruning the tokens $\{x_{j, j\in \psi'}\}$, with $\psi_0' \subseteq \{1,...,n\}$:
\begin{equation}
\mathbf{Y}'_0=\sum_{i=1}^n \mathcal{A}_{1,i} \mathbf{V}_i - \sum_{j \in \psi_0'} \mathcal{A}_{1,j} \mathbf{V}_j = \mathcal{A'} \mathbf{V'}.
\end{equation}
Replacing $y_1$ by the original $\mathit{cls}$ token $\mathbf{Y}$, while replacing $y_2$ by the pruned $\mathit{cls}$ token $\mathbf{Y}'_0$ in Eq.~(9), if we have
\begin{equation} \label{eq:a}
    {
    \Vert \mathcal{A} \mathbf{V} - \mathcal{A'} \mathbf{V'}  \Vert_2 < \epsilon,
    }
\end{equation}
then with the prior conclusion from Eq.~(9)
\begin{equation} \label{eq:b}
    {
    \Vert \mathcal{I}(\mathbf{Y}) - \mathcal{I}(\mathbf{Y}'_0) \Vert = \Vert \mathcal{I}(\mathcal{A} \mathbf{V}) - \mathcal{I}(\mathcal{A'} \mathbf{V'}) \Vert \ \leq \ \eta. 
    }
\end{equation} 
To maintain information, we choose $\eta_0 = \eta' << 1$. Supposing that $\epsilon$ associated with $\eta'$ is $\epsilon'$, then according to Eq.\,(\ref{eq:a}) and Eq.\,(\ref{eq:b}), for a subset $\psi'$ such that
\begin{equation} \label{eq:12}
    {
    \Vert \sum_{j \in \psi'} \mathcal{A}_{1,j} \mathbf{V}_j \Vert_2 < \epsilon',
    }
\end{equation}
we can approximate the informativity of $\mathit{cls}$ token as:
\begin{equation} \label{eq:13}
    {
    \mathcal{I}(\mathbf{Y}) \approx \mathcal{I}(\sum_{i=1}^n \mathcal{A}_{1,i} \mathbf{V}_i - \sum_{j \in \psi'} \mathcal{A}_{1,j} \mathbf{V}_j) = \mathcal{I}(\mathbf{Y}').
    }
\end{equation}
Thus, Eq.~(11) is proved. Such analysis indicates, pruning/merging tokens with sufficiently small semantic value, barely affects the informativity of the $\mathit{cls}$ token.

\begin{figure*}[t]
\begin{center}
\includegraphics[width=0.99\textwidth] {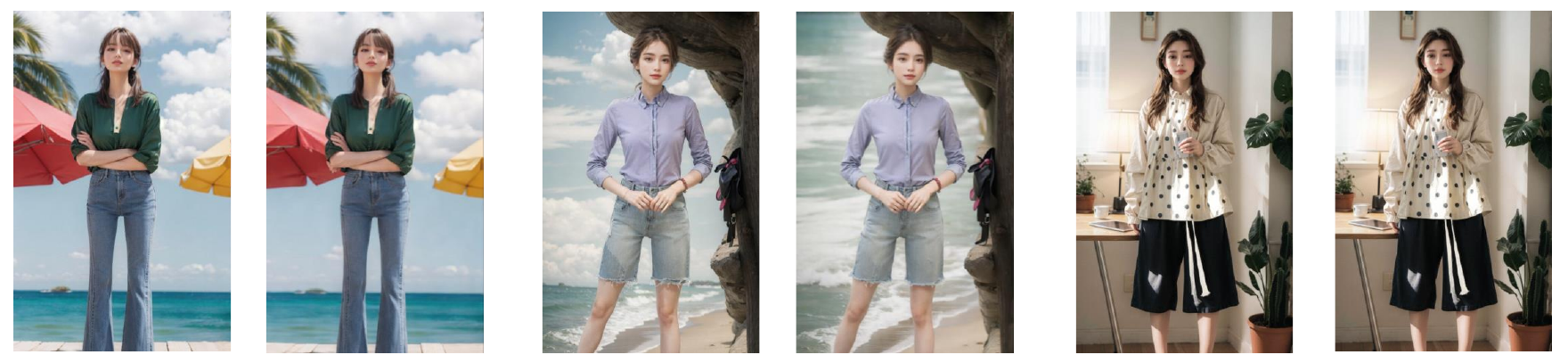}
\end{center}
\caption{\textbf{Visualizations of Text-to-Image Generation.} Left is no acceleration, and Right is acceleration by our Turbo module.}
\label{fig:textgen}
\end{figure*}

\begin{figure*}[t]
\begin{center}
\includegraphics[width=0.99\textwidth] {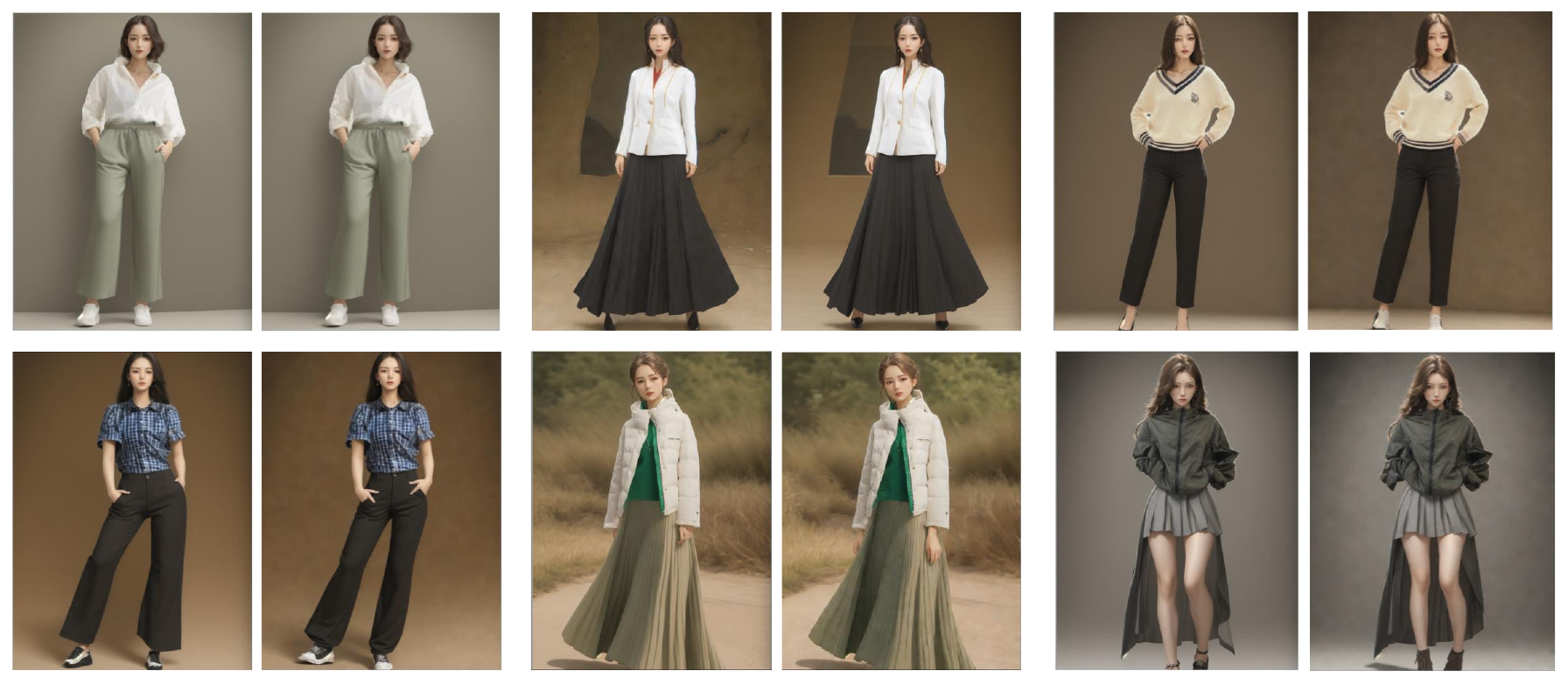}
\end{center}
\caption{\textbf{Visualizations of Image-to-Image Generation.} Left/Right is no/Turbo acceleration. The generation quality is very similar before and after acceleration.}
\label{fig:imgsgen}
\end{figure*}

\section{Visualization Results}
To intuitively demonstrate the superiority of our Turbo module, we here visualizes results from generative tasks. Figure~\ref{fig:textgen} shows the results of text-to-image generation. The used text prompts are usually as: One female model wearing purple long sleeves and blue jeans stands on the coast. Although Stable Diffusion still needs improvement in generation details (such as hands), Turbo acceleration has almost no side effects. Figure~\ref{fig:imgsgen} shows the results of image-to-image generation. Generally speaking, the conclusion is similar: the generation quality is close before and after Turbo acceleration.

\bibliographystyle{splncs04}
\bibliography{egbib}
\end{document}